%% file: manuscript.tex
\documentclass[sigconf]{acmart}

\usepackage{booktabs,tabularx,stfloats,microtype} 
\usepackage{float}
\AtBeginDocument{%
  \providecommand\BibTeX{{%
    \normalfont B\kern-0.5em{\scshape i\kern-0.25em b}\kern-0.8em\TeX}}}

\copyrightyear{2022} 
\acmYear{2022} 
\setcopyright{acmcopyright}
\acmConference[IDC '22]{Interaction Design and Children}{June 27--30, 2022}{Braga, Portugal}
\acmBooktitle{Interaction Design and Children (IDC '22), June 27--30, 2022, Braga, Portugal}
\acmPrice{15.00}
\acmDOI{10.1145/3501712.3529747}
\acmISBN{978-1-4503-9197-9/22/06}

\acmSubmissionID{3383}

\begin{document}

\title{Understanding Factors that Shape Children's Long Term \\ Engagement with an In-Home Learning Companion Robot}


\author{Bengisu Cagiltay$^*$, Nathan White$^*$, Rabia Ibtasar$^{**}$, Bilge Mutlu$^*$, and Joseph Michaelis$^{**}$}
\affiliation{%
  \institution{$^*$Department of Computer Sciences, University of Wisconsin--Madison}
  \country{Madison, Wisconsin, USA}}
\affiliation{%
  \institution{$^{**}$Learning Sciences Research Institute, University of Illinois Chicago}
  \country{Chicago, Illinois, USA}}
\email{bengisu@cs.wisc.edu, ntwhite@wisc.edu, ribtas2@uic.edu, bilge@cs.wisc.edu, jmich@uic.edu}


\renewcommand{\shortauthors}{Bengisu Cagiltay, Nathan White, Rabia Ibtasar, Bilge Mutlu, and Joseph Michaelis}

\begin{abstract}
Social robots are emerging as learning companions for children, and research shows that they facilitate the development of interest and learning even through brief interactions. However, little is known about how such technologies might support these goals in authentic environments over long-term periods of use and interaction. We designed a learning companion robot capable of supporting children reading popular-science books by expressing social and informational commentaries. We deployed the robot in homes of 14 families with children aged 10--12 for four weeks during the summer. Our analysis revealed critical factors that affected children's long-term engagement and adoption of the robot, including external factors such as vacations, family visits, and extracurricular activities; family/parental involvement; and children's individual interests. We present four in-depth cases that illustrate these factors and demonstrate their impact on children's reading experiences and discuss the implications of our findings for robot design.
\end{abstract}



\maketitle

\section{Introduction}
\input{teaser}
Understanding how to sustain long-term engagement of technology has been a great challenge in human computer interaction (HCI), perhaps even more so for educational technologies that target children \cite{serholt2016robots}. Designing technology for long-term use in home environments has unique challenges due to the dynamic and complex nature of family relationships, routines, and habits \cite{beech2004lifestyles,neustaedter2009calendar,davidoff2010routine}. Moreover, the different use cases and needs of each family lead to further challenges for establishing a diverse understanding of the design requirements.
As social and educational robots increasingly become commonplace in home environments, there is growing interest among human-robot interaction researchers to better understand the factors that affect robot adoption and long-term interactions \cite{irfan2019personalization}. However, there is much to uncover for the design of social robots to support these more complex and dynamic interactions that are unique to the home environment.


Design features such as personalization \cite{irfan2019personalization}, inclusion of meaningful activities \cite{coninx2016towards}, and supporting interaction in a family environment \cite{cagiltay2020investigating} can help facilitate child-robot relationships that are sustained in the long-term. To serve as a context for long-term interaction in a home environment, we have designed a learning companion robot that reads with children. Prior work has shown that a reading companion robot can support children's learning \cite{michaelis2019supporting}, comprehension \cite{yueh2020reading}, and engagement in the reading activity \cite{michaelis2018reading}. 
While there exists compelling evidence for reading to be used as an activity to facilitate long-term engagement and relationship, additional consideration is needed for the design of robots and activities in order to address the unique challenges that stem from long-term interaction \cite{leite2012modelling}.
Therefore, in order to sustain long-term robot interactions, there is a need to observe the use of robot systems in home environments to better understand the factors that facilitate engagement, as well as factors that result in disengagement and abandonment. 

While we do know many reasons regarding user abandonment of robot technology \cite{de2017they} or breakdowns\cite{serholt2018breakdowns}, little is known about how the complex and dynamic nature of day-to-day in-home life will affect the interactions children have with social robot over an extended period of time. Many external factors outside of the design of the robot and the activity can affect sustained use. These factors range from other commitments such as sports practices, hobbies, chores, vacations, and family obligations to individual and family context including routines, values, beliefs, and even illnesses, all of which could be competing for the child's limited time and attention. These factors, which can cause extended breaks between interactions, can negatively affect a user's interest or desire in continuing to use the technology. Therefore, there is a need to identify what these external factors are and to understand what impact they have on children's long-term interactions and use of robots.

In this paper, we aim to address this gap by investigating how children use a learning companion robot over the course of a four-week in-home deployment and how the robot integrates within their daily life. To this end, we designed a robot that interacts with children over informal science reading, providing social and informative commentary as the child reads a book aloud to the robot.
We deployed this robot in a naturalistic uncontrolled user study for four weeks during the summer. Through an analysis of the experiences of children interacting with this robot, we seek to answer the following research question (RQ): \textit{What are the factors in authentic in-home child-robot interactions that affect children's interest in reading popular science books and long-term engagement with a learning companion robot?}

\section{Related Work}
Storytelling and reading aloud to either peers or younger children, or being read to by parents or teachers is a common practice in households, schools, and community centers \cite{barton1986tell}. This practice of reading aloud is an educational activity that helps children with word pronunciation, reading comprehension, attention to detail, and their ability to make connections between their personal life and the story \cite{duursma2008reading, barton1986tell}. In educational settings, reading aloud supports confidence, independence, and enjoyment of children who struggle with reading along, comprehension, and success in class \cite{dreher2003novel}. Particularly for in home environments, the shared activity of reading aloud can help with young children's social-emotional development and support their relationships with parents or family members \cite{mendelsohn2018reading}. However, factors that affect children's motivation to read is multidimensional, impacted by intrinsic or extrinsic motivations and goals, their self-efficacy, or other social aspects \cite{wigfield1997relations}, and their long-term motivation in reading can be predicted from their situational interest towards a book \cite{guthrie2005spark}. Technology designed to support children's reading motivation and their informal learning practices at home carries a great importance in supporting children's intellectual and social development.

Prior research has shown the usefulness of social robots and their beneficial impact on children's education and learning habits. Social robots designed for school environments can support social interactions \cite{kanda2007two}, or topics such as math \cite{brown2013engaging, lopez2018robotic}, science \cite{shiomi2015can, davison2020working}, or second language learning \cite{westlund2016tega, gordon2016affective, kennedy2016social}. A number of social robots have also been introduced into the home environment, aiming to supplement children's in-home learning \cite{han2005educational, michaelis2019supporting} or provide social interventions for children with autism \cite{clabaugh2019long, scassellati2018improving}. \citet{han2005educational} found that the use of in-home educational robots can better support a child's learning concentration and interest, as well their academic achievement in learning the English language, compared to web-based instruction or books accompanied by audio. \citet{michaelis2019supporting} found that after two-weeks of reading with a robot at home, children had increased accuracy in learning assessments and that the robot had supported their situational interest. \citet{scassellati2018improving} found that, after a one-month in-home intervention with a social robot, children with autism were observed to have significant improvements in their social skills including making eye-contact, and both initiating and responding within conversations. Similarly, \citet{clabaugh2019long} found that autonomous and personalized interventions during an in-home and long-term socially assistive robot deployment for children with autism were effective in supporting children's math skills.



In these in-home environments, interactions are governed completely by the users, giving them more freedom of control and agency as they decide the when or how of the interaction. This opposes to more structured and scheduled activities in school or lab environments, where we know that robots can be successful in sustaining interaction over extended periods of time \cite{kanda2007two, davison2020working} as well as positively influencing children's education \cite{park2019model, kory2019long}. For example, \citet{kanda2007two} deployed a robot in an elementary school for two months, allowing children to interact with the robot during their recess period. \citet{davison2020working} deployed an unsupervised robot into the classroom for four months, allowing children to interact with it voluntarily whenever they wanted to. These setups mimic the in-home environment most closely by allowing children to have more control over the interaction.

However, a major challenge for all in-home social robots is the maintenance of long-term interactions. Few studies have demonstrated sustained human-robot interactions in the long term \cite{kanda2007two, leite2013social, scassellati2018improving, de_graaf_why_2019}.
These long term in-home studies have uncovered major challenges for sustaining user engagement, most commonly that many users eventually decrease technology usage or even stop using it altogether \cite{de2016long}. To understand why this abandonment occurs, \citet{de2017they} explored the reasons why adults would stop using robots in the home over the course of a six month study and categorized them into resistors, rejectors, and discontinuers as described by \citet{wyatt2002they} and \citet{rogers2010diffusion}. De Graaf et al. found that each category had different reasons for abandoning the robot; i.e., language and privacy concerns for resistors, disenchantment and the restrictions and problems with interactions for rejectors, and finally system adaptability and sociability for discontinuers. To address similar abandonment issues with children, possible solutions such as designing adaptive robots that are capable of switching between multiple activities was found to positively support children's perceptions about the robot \cite{coninx2016towards}. However, robots should adapt in many ways to compliment family routines, which is crucial for ensuring the integration of social robots into the complex daily life of children and families. 

Overall, in-home environments are complex and dynamic. It is necessary to explore how factors such as family routines, relationships, dynamics, and values in the home can affect long-term child-robot interactions. Our work aims to address the gap of understanding the complexities of children's long-term engagement in authentic home environments with a learning companion robot.

\section{Method}
\subsection{Resources}
We delivered 1) a robot and its equipment 2) reading supplies and 3) bi-weekly newsletters to the participating families\footnote{All resources and robot interaction design guidelines described in this paper are shared on OSF \url{https://osf.io/bks8w/?view_only=21099ef7102e4216a087bc8e117b2c75}}.

\paragraph{Robot and Equipment}
We used a Misty II robot \footnote{\url{https://www.mistyrobotics.com/}} due to its compactness for use in in-home deployments and semi-humanoid design. Misty provides a 4-inch LCD display as its face, allowing for highly customizable facial expressions and animations. We provided other equipment including a Raspberry PI 4, a mobile hotspot, and a multi-outlet power extension cord. The Raspberry PI acted as the main computing power for the setup by reacting to the different sensory inputs of the robot and sending commands back. This connection between the robot and Raspberry PI was facilitated through the mobile hotspot, which also allowed us to upload study data nightly and accept potential program updates. Updates sent throughout the study did not affect the robot's behavior, but instead fixed issues that prevented the designed interaction from taking place, such as the Raspberry PI crashing when briefly losing connection to the robot.

\paragraph{Reading and Interaction Supplies} 
The supplies provided to the families include: 20 informal science books, a reading journal, a robot interaction tutorial booklet, topic cards for the tutorial interaction, a card for guest use, four cards for volume control, and two informational cheat sheets. The \textit{books} were selected to motivate readers \cite{hidi2000motivating} and promote deeper learning \cite{renninger2015power}. We curated a varied selection of 20 books by considering different factors such as book length, lexile scores, book topic, and book style that a range of children may find appealing.
The \textit{reading journal} and \textit{guest card} allowed for the robot to identify who it was reading with and address them by name. The \textit{journal} also allowed children to keep track of their daily reading and express how they felt about the reading. \textit{Volume cards} allowed users to set the robot's audio level. In addition, children received a \textit{tutorial booklet} describing and demonstrating the robot's functionalities. The \textit{cheat sheets} had information about the list of resources provided, how to interact with and care for the Misty robot, answers to frequently asked questions, and how to travel with Misty.

\paragraph{Bi-Weekly Newsletter} During the second and fourth weeks of the study, we mailed newsletters to each family. These newsletters contained three sections: (1) Science Article, (2) Community Connections, and (3) Family Science Activity. The newsletters were designed based on the themes, ``Space'' and ``Nature and Pollution.'' The ``Science Article'' section included a short paragraph from a selected article from frontiers-kids\footnote{\url{https://kids.frontiersin.org}}. The ``Community Connections'' section included a summary of a selected book and all participants' top three most read books so far with the robot. The ``Family Science Activity'' section included a summary of a simple science activity and a link to webpage with detailed instructions. %

\subsection{Interaction Design}
\subsubsection{Interacting with the Robot}
Users interacted with the robot by showing reading materials that are equipped with April Tags\footnote{\url{https://april.eecs.umich.edu/software/apriltag}} and pressing sensors on the robot. The robot could understand which page is read and respond with a social commentary by detecting the April Tags that were attached to every 3-4 pages of the books. Users could also inform the robot about their actions through pressing one of four corner bumper sensors.
The \textit{Repeat} bumper repeats the most recent comment from the book. The \textit{Pause} bumper pauses the reading interaction, allowing users to take a brief break up to 10 minutes. 
The \textit{Yes/Continue} bumper allows the user to confirm their actions when prompted by the robot or continue the interaction from the paused state.
The \textit{No/Quit} bumper allows the user to end the interaction or respond to the robot if prompted.

\subsubsection{Robot's Social Commentary}

\paragraph{Knowledge Support Comments}
These comments were created to leverage the science content in the books. They were designed to strategically emphasize the science content and establish connections between the child and book characters doing science. These supports were divided into four categories based on previous research on supporting children's reading and comprehension while reading books \cite{ivey2000tailoring}. The robot's \textit{prediction comments} shared a valid prediction about science content in the book based on recent relevant information/clues from the text. \textit{Summarize comments} identified important information, themes, and ideas within a text and communicated these in a clear and concise statements to capture the essence of the text. \textit{Questioning comments} focused on the basic understanding of materials especially when considering complex scientific phenomenon. Finally, \textit{vocabulary supports} provided clarification comments for difficult words and scientific terms as comprehension of such words has been known to support student's knowledge building.
\input{fig_interactionflow}

\paragraph{Social Support Comments}
From an educational perspective, social support \cite{leite2012modelling}, different levels of social behaviors \cite{kennedy2015comparing}, and personalization \cite{gordon2016expanding, leyzberg2014personalizing} can facilitate a positive interaction between children and robots. The social comments were designed as a type of social support for children's learning during this child-robot reading activity. Four categories of social supports were created based on activities that were shown to be beneficial through previous research \cite{kanda2009affective, prendinger2005empathic, gockley2005designing}. \textit{Robot self-disclosure} comments helped children to identify the robot as a peer and friend by disclosing personal information and beliefs. \textit{Recall past interactions} comments made references to prior shared activities between the child and the robot. These included references to previous books the child may have read with the robot or an earlier event that was mentioned in the book currently being read by the child. Such comments are a way to create a shared history with the child and build stronger social connections during this reading activity. \textit{Memory and adaptation} comments allowed us to build a personalized experience for every child by embedding information about them in the robot’s responses such as the child’s name and their topic preferences. These comments connected with the child by building on their topical interest as prior work has shown this can be especially motivating during reading activities for students \cite{hidi2000motivating}. Finally, the \textit{emotional response} comments explicitly express the emotional state of the robot through specific visual and audio responses that are accompanied with verbal responses. Together, these social support comments help to build the robot's character and personality to facilitate the child’s social bonding with the robot during the reading activity.

\paragraph{Interest Comments}
Interest comments were tailored to meet the specific interest-development needs of the profile of children targeted in the study. These interest supports were divided into two categories based on findings from previous work \cite{durik2015if}: value and belongingness. The \textit{value comments} conveyed the value of the theme of the book by relating the theme to something that would be of importance or valuable to children, their families, and their communities and/or society at large. \textit{Belongingness comments} choose elements in the books to highlight the work, skills, and practices of scientists. These comments draw attention to activities that children can do that are similar to what scientists do and how they are doing it in the books currently being read.

\subsubsection{Interaction Flow}
Children experienced two types of interactions with the robot. The first-time interaction which occurred once as a tutorial, and the regular reading interaction which occurred each time the user turned the robot on. 
For the \textit{first-time interaction}, children started reading the tutorial booklet, became familiarized with interacting with the robot, its functionalities, and the flow of the reading activity. As part of the tutorial, children rated different book topics from most to least liked, allowing the robot to recommend books later on. Researchers were also present to provide guidance as needed. 
For the \textit{reading interaction} as illustrated in Fig. \ref{fig:flow}, users first turn the robot on and see a sleeping robot face while the system boots up. Once fully booted, the robot prompts the user to scan their reading journal to identify the reader. Following this, the robot greets the user by name and expresses excitement for reading with them. The robot then asks the user to either continue reading their previous book or select a new one. When suggesting new books, the system will look at the user's ranking for different book topics and their lexile score to narrow down the book selection to prioritize books the user will enjoy. After the user selects a book, the robot suggests a reading goal for the day based on their average reading time. The robot then expresses excitement telling the user they can begin reading. During the \textit{reading phase}, users read the book aloud to the robot and show the April Tags as they come across them to prompt the robot for comments. If the user finishes the book, the robot will ask them if they enjoyed it, and will update the user's topic preferences.  
Once the user completes their desired time of reading for the day, they press the ``quit'' bumper to terminate the interaction and are prompted to place the robot on its charging pad.  


\subsection{Procedure}
The study included two stages, a one-week baseline phase (without the robot), and a four-week robot interaction phase. 
During the \textit{baseline} week, children read the provided books or their own books and logged their daily reading habits in their reading journal. This week served as a baseline of reference for children's regular reading habits without the robot.
During the four-week \textit{robot interaction} phase, children read aloud the provided books to the robot at their own pace, schedule, and liking. Only on the first day with the robot, the researchers were present to remotely help setup the robot and children started the first-time interaction, completed the tutorial, and read aloud to the robot for 10-20 minutes. After the first-time interaction, children interacted with the robot on their own, but attended weekly meetings with the researchers including a semi-structured interview about their experience and thoughts about the robot that lasted around 20-30 minutes. 

\subsection{Measures} 

\subsubsection{Semi-Structured Interviews}
In total, we conducted seven semi-structured interviews with the participating families. The first interview focused on the child's reading habits and the family's ways of motivating to read and was administered on the first day of the study's baseline week. The second interview was administered the next week, before the child met the robot, and focused on the child's experiences during the baseline week. The third interview was administered right after meeting the robot and completing the tutorial interaction, and focused on the child's initial impressions and experiences with the robot. The remaining four interviews were administered weekly during the robot interaction weeks, focusing on the child's and family's experiences with the robot. 

\subsubsection{Interaction Logs from Reading with the Robot}
Every family's interaction with the robot was logged, including the following information: date and time of the interaction, books read, reading duration, and a chronological list of events that occurred in the interaction. Examples of the types of events included are the robot's speech, the type of input the user gives, and the type of shutdown that occurs, i.e., system crash, robot is turned off, or successful shutdown.

\subsection{Participants}
We recruited 16 families 
with children aged 10--12 through email solicitation, distributed by local community center staff as well as through the use of institutional staff mailing lists. 14 children (6 boys, 8 girls; mean age 11.4) completed the full study duration, and two families ended the study early due to technical failures with the system. The recruitment criteria included children that have low interest in science and were between the ages 10--12 at the time of sign up. We identified children with low interest in science through a pre-screening questionnaire sent to the parents. Non-eligible siblings were allowed to participate and interact with the robot, although their data was not included in the analysis. Each family received \$50 as compensation after completing the study.

\subsection{Analysis}
We conducted a Thematic Analysis following the guidelines of \citet{clarke2014thematic} to identify the factors that shaped children's long-term engagement with the robot. The first author was familiarized with the data through conducting all interviews with the participants. The interviews were transcribed via an online auto-transcription service and were manually reviewed by the first author for correctness. The interaction logs were extracted, formatted, and analyzed by the first and second authors. The first author generated a codebook and reviewed the codes and themes with the remainder of authors until reaching an agreement and reported .





\input{fig_results}

\section{Results}
The results of our thematic analysis indicated many idiosyncratic experiences for children during the long-term study. To capture those experiences we identified emerging themes around two major areas: (1) critical factors that influenced the long-term interaction with the robot, and (2) the effects of those factors on children's experiences reading with the robot. First, in Section \S\ref{subsec:Factors}, we describe the critical factors we identified that appear to have strongly influenced the children's experience and engagement with the robot. 
In Section \S\ref{subsec:CaseStudies}, we present four cases to illustrate \textit{how} these factors shaped children's long-term robot engagement (See Figure \ref{fig:results}). We organize our cases by thematic structure of types of experiences with the robot, and emphasize specific factors that influenced those experiences. To protect the privacy of the children and families that participated in our study, we replaced children's real names with aliases. 

\subsection{Factors That Influenced Children’s Long-Term Engagement With the Robot}\label{subsec:Factors} 




\paragraph{\textbf{Life Events and Interruptions}}
The minimal control on the interaction and naturalistic structure of our study design allowed for families to continue their regular routines and lifestyles but varied the frequency of robot interactions (See Table \ref{tab:usage}). Life events including week-long vacations, short family visits, daily sports practices, national holidays, new hobbies or classes, etc. were external factors within our study that have affected children's ability to sustain regular interactions with the robot. Three families had week-long vacations during the study, while many other families had shorter trips, including 1-2 day weekday/weekend trips or friends and family visits. These short and long term interruptions impacted children's engagement differently throughout the study, mostly leading children to focus less on interacting with the robot and occasionally skipping one of the weekly meetings with the study team. 

\paragraph{\textbf{External Prompts as Motivational Factors}}
Parental involvement acted as an external prompt that affected children's long term engagement. While family members' attendance to the meetings were encouraged, it was not required. Most of the children preferred to attend the meetings on their own and five of the children's parents or family members regularly joined the weekly interviews. Regularly attending \textit{family members' involvement in the weekly meetings} allowed them to share personal opinions about the experience or motivate the child to share forgotten details about the interactions with the robot, e.g., not reading aloud, experiencing technical issues, or an exciting moment with the robot. 
\textit{Parental involvement at home} affected children's engagement as it was a source of reminders or prompts to read with the robot. During the interviews, some children expressed self-motivation about initiating an interaction with the robot and did not require their parents to prompt them, while others shared that their parents would remind them occasionally to read. However, over time, two parents expressed observing changes in their children's motivation to read with the robot which also influenced a change in their level involvement. Parents usually described this as a decrease in the novelty of the interaction which had impacted the child's self-motivation, thus requiring them to prompt their children more often to promote engagement. This change influenced children's perspectives towards reading with the robot, shifting from a fun activity to a chore or an item on a to-do list. Children from families that relegated the activity to a daily to-do list expressed that this took away the excitement from the activity. 




\begin{table*}[!b]
    \caption{Total Usage Frequency and Reading Duration Over Time (adapted from \citet{de2016long})}
    \label{tab:usage}
     \centering
    \begin{tabularx}{\textwidth}{lccccc}
         \toprule
 \textbf{Usage} &\textbf{Baseline}&\textbf{W1}&\textbf{W2}&\textbf{W3}&\textbf{W4} \\
         \midrule
 \textit{Frequency} &&&&\\
 Never &3&0&1&2&4\\
 Once a week &1&1&2&4&3\\
 A few times a week &10&10&11&7&7\\
 Once per day &0&0&0&1&0\\
 At least once per day &0&3&0&0&0\\
 \hline
 \textit{Duration at a time} &&&&&\\
 Less than 5 minutes &3&0&1&3&4\\
 5 to 10 minutes &1&1&0&2&0\\
 10 to 15 minutes &1&7&5&1&1\\
 15 to 20 minutes &3&4&4&5&6\\
 More than 20 minutes &6&2&4&3&3\\
          \bottomrule
Baseline: 1-week data from reading journal entries prior to meeting the robot \\W1,W2,W3,W4: Weekly data from interaction logs reading with the robot\\
    \end{tabularx}
\end{table*}
\paragraph{\textbf{Immediacy, Routines, and Self Motivational Factors}}
Many children mentioned that having the robot or reading materials placed in an immediate location that is highly visible, such as the common area in their living room, their bedroom, or their play/study room, helped them with noticing the robot and being motivated to start reading with it. Children that placed the robot and reading materials in a immediate location mentioned they were able to fit the robot into their daily routines more easily. Reading with the robot was most commonly a part of the bedtime routine, or the morning routine. However, in two cases children lost the factor of immediacy due to moving the robot or resources to a different location. This negatively affected their engagement as the robot was no longer in a place they would frequently see it and limited the visual reminder for engaging in the task. Moreover, a lack of alternative entertaining activities at home and having the robot as a ``quick fun activity to do'' also motivated some kids to read, as expressed by three of them. 

Some children had unique self motivation factors that impacted their reading and engagement with the robot. For example, one child had a routine to read every day at a scheduled time, except she took a fixed day off from reading with the robot every week. Furthermore, three kids shared that a partial motivation for them was receiving compensation at the end of the study. Two children described that they would use the compensation to add to their savings or purchase a gift for themselves at the end of the study.

\paragraph{\textbf{Individual Interest in the Robot's Interaction Flow}}
Children's experiences with the robot's interaction flow (shown in Figure \ref{fig:flow}) affected their long term engagement. While children generally liked the flow of the interaction between these phases at the beginning of the study, at later weeks few children had diverse opinions. Three children expressed the flow being repetitive while others still felt excitement during the \textit{wake-up} and \textit{identification} phase. In \textit{book recommendation phase} the robot's personalized book suggestions were enjoyed by children and often factored into children's book selections. During the \textit{reading phase}, the robot's idle movements and the activities of reading aloud, showing the tags on the book's pages, and listening to the robot's comments were perceived differently by children. 

The robot's \textit{idle movements} were described to be non-distracting by children, but occasionally the robot's random movements were perceived as confusion, or as a reaction to an event happening in the environment or in the book. Eight children expressed that \textit{reading aloud} helped with understanding the details about the reading, pronouncing words, and felt like they were reading to a younger child without the interruptions. Over time, some children felt it slowed down their reading, felt tiring, and in some cases made them feel self-conscious. For these reasons, three children shared that they occasionally stopped reading aloud to the robot. The experience of \textit{showing the tags} and \textit{listening to the robot's comments} was mostly perceived as a pleasant break from reading aloud or as an activity that created a connection between the child and robot by learning about the robot's opinions. Diversely, four children expressed that showing the tags to the robot was a way to keep track of the pages read, like a bookmark, and two children ignored the tags unless it was the last page they read for the day. In general, more than half of the families found the experience of reading to a robot similar to reading to a dog/pet, as the robot did not feel judgmental, did not interrupt them, and made them feel like it was actually listening and attending to them.

\paragraph{\textbf{Individual Interest in the Robot's Social Commentary}}
The robot's social commentary and expressions during the reading phase was a factor that impacted children's engagement with the robot. While children mostly enjoyed the robot's verbal and non-verbal expressions, their perspective on some of the robot's commentary types, including knowledge support (prediction, summarize), social support (self-disclosure, emotional response), and interest support (value) commentaries varied throughout the study.
The robot's \textit{prediction} comments were found to spark curiosity about the book. The robot's \textit{summarize} comments helped children understand details by pointing them to important sections. A small set of children expressed that they found these informative and book related commentaries uninteresting and repetitive. 
The robot's \textit{self-disclosure} commentaries were interesting to the children as they felt the robot was more human-like than they expected and made them excited to get to know the robot more closely, creating a sense of connection making. Nearly half of the children excitedly shared such comments where ``Misty went on a road trip'', or where ``Misty talked about a science experiment she did with friend.'' Nearly all children enjoyed the robot's \textit{emotional response} comments and how the robot transitioned between emotions during the reading phase. For some, these emotion expressions were a pleasant surprise that made the robot more human-like and motivated them to read more. More than half of the children expressed they were excited to share the robot's various emotions that they have discovered with their families or they shared it with the experimenter as the first topic during the interviews.
Children had mixed feelings about the \textit{value type} comments. Children described these comments as ``the comments about me'' or ``compliment comments.'' Four children expressed they liked the compliments as it felt like the robot supported and believed in them, however for two children it felt irrelevant in situations if it was regarding a topic they were not interested or good in. For example, one child mentioned not being good in math, and when the robot shared a comment that valued her math skills, it did not relate to her. For nearly half of the children, the \textit{value} comments started to feel repetitive or ``too positive'' over time after their novelty passed.

Conversely, there were a few children for which the robot's commentaries did not carry any significance at all, or they were not interested in the robot's opinions. When asked about their experiences, these children mostly responded indifferently by saying they do not listen, know, remember, or care about the robot's comments.


\input{fig_overallreading}

\paragraph{\textbf{Individual Interest in the Materials}}
Children's interest in the books changed over the study, affecting their engagement with the robot. Among the provided 20-books, which included a variety of topics and genres, five books from the set weren't read by any child and many children did not complete the books they have started throughout the study (see Figure \ref{fig:booksread}). Some children quickly read the books they were interested in during the first or second week, and in the later weeks had difficulty expanding their selection which affected their motivation to keep reading. As a solution, nearly all children tried new genres of books they hadn't tried before, which helped with sustaining or increasing their engagement with the robot. Conversely, others were hesitant on expanding their book selection which resulted in a decrease of engagement with the robot. As the books covered a range of reading levels, rarely some children could not advance in a book due to it being too difficult for them even if they found the topic interesting. 

The novelty of the bi-weekly newsletters also sparked the interest of some children throughout the study and even influenced them to read new books. In total, three children read the first newsletter and four children read the second newsletter with the robot as a regular book, while the rest expressed that they read it on their own, without the robot. Five children mentioned that the top read books in the community connections section of the newsletter inspired them to try out new books because they were curious to see what other children were reading.

\subsection{Cases Demonstrating How Factors Shaped Children’s Long-Term Engagement With the Robot}\label{subsec:CaseStudies}
Having established the critical factors that influences child long-term engagement with the companion robot, we now turn to presenting four cases we identified that illustrates \textit{how} these factors were influential in nuanced ways (See Figure \ref{fig:results}). Each case was chosen by grouping children that exhibited similar experiences in long-term engagement with the robot, resulting in four categories: 1) Children that Modified their Interaction 2) Children that Discontinued the Use of the Technology 3) Children that were Interrupted 4) Children that Adopted the Technology. 

\subsubsection{\textbf{Case 1: Children that Modified their Interaction}} Three children adjusted their interaction with the robot over time, by changing their reading style by \textit{week three}. The recommended interaction with the robot was to read aloud and show the AprilTags on the book's pages when available. Children in Case 1 modified their interaction style based on their preferences, e.g., initiated the interaction with the robot but did not read aloud or did not show the tags to the robot. The critical factors influencing these children were changes in their individual interest in reading aloud, the robot's social commentaries and expressions, the books, and the external prompts and reminders from parents.

For example, \textbf{Stacey} (12yo, female) enjoyed the books and the robot's commentaries but stopped reading aloud to the robot in week three. This was explained by her parents as, \textit{``there are times when she's not reading aloud''} while continuing to show the tags and receiving social comments. Stacey explained, \textit{``I didn't get bored. I like the book and I like her[Misty's] comments.''} Murphy (12yo, male) and Craig (10yo, male) both stopped reading aloud to the robot on week three and also stopped showing the tags to the robot, expressing low interest towards the interaction and the robot's comments. Murphy's interaction logs showed that he read once a week, mostly on the interview days with the experimenter, suggesting the weekly interviews factored into Murphy's motivation to read with the robot. \textbf{Murphy} expressed his low interest in the robot's commentaries as \textit{``well, I don't care, her[Misty's] comments don’t really make an impression on me.''} \textbf{Craig}'s interaction logs, however, showed that he read with the robot daily but the interviews reveled that ``it's definitely challenging to find books to read (...) some of them[books] are really long and they just don't interest me.''  

\subsubsection{\textbf{Case 2: Children that Discontinued the use of Technology}}
Two children discontinued the use of the robot after the second week due to critical factors that relate to changes in their interest for the robot's interaction, social commentary, reading aloud, or the available materials to read.

For example, \textbf{Penny} (11yo, female), enjoyed the robot's interaction however had depleted books that she found interesting. As pointed out by her parent, \textit{``finding books is Penny's issue ... she reached her capacity for the books,''} which led to discontinued interactions with the robot. 
\textbf{Allen} (12yo, male), enjoyed the books, however did not enjoy reading aloud as \textit{``it takes longer to read with Misty''}, which led to a discontinuation in reading with the robot but he continued reading the materials on his own. Allen's parent described this change as, \textit{``they would just do it, and then it slowly got to the point of I had to remind them for, and they forgot or we were busy, and none of us remembered. We got to a point where I was reminding them on a regular basis, and then it seemed like it was becoming more of a chore than it was the excitement.''} 


\subsubsection{\textbf{Case 3: Children that were Interrupted}}
Long-term interruptions during the study, including vacations, family visits, national holidays, hobbies affected children's engagement with the robot. Case 3 includes three children that experienced at least one week-long interruption and cancelled at least one weekly interview.

The experiences of children in this case differed based on the nature of the interruption and their individual attitudes towards the robot. For example, Naomi's (11yo, female) and Jimmy's (12yo, male) experience suggest that even though they both had a week-long interruption due to family vacations, their positive attitude towards the robot and the reading materials supported them to continue interacting with the robot. \textbf{Naomi}'s parent described her interest as \textit{``Naomi has a pretty strong impulse for caring. Having that interaction with a life-like thing was something that was interesting to her.''} Although she did not enjoy reading aloud, she still put in the effort, saying \textit{``somebody's listening so I have to try and do my best because I'm pronouncing words and speaking loudly.''} Similarly, \textbf{Jimmy} enjoyed the robot's companionship and full attention while reading \textit{``I like how it felt like a reading buddy. Because I feel less rushed. Because sometimes if I'm reading to someone, I have limited time. But it(the robot) is not going to go do an errand or something.''} On the other hand, \textbf{Chuck} (12yo, male) had the most frequent amount of interruptions, as he visited close relatives three days a week and was on a family vacation during his third week. Chuck also expressed low interest in the robot's comments and found it challenging to find a factor that motivated him to read. The combination of these long-term interruptions and low interest in the robot and reading made it challenging for Chuck to regularly interact with the robot.



\subsubsection{\textbf{Case 4: Children that Adopted the Technology}}
Six children adapted the technology and interacted with the robot regularly on a daily or almost daily basis. However, the motivational factors for each child in this category varies from the robot's interaction design and external factors. A combination of critical factors shaped how children maintained their engagement with the robot, including having individual interest in the reading materials, the robot's interaction and social commentary, parental prompts to read, the robot's immediacy, and the child's routine. 

Here we share some example quotes demonstrating how the critical factors shaped children's engagement and perceptions in this case. For example, the robot's expressions shaped \textbf{Pamela}'s (12yo, female) perceptions of the robot, \textit{``she's super cute like I love her faces change and everything, and I can tell that she's really listening.''} 
\textbf{Alice} (12yo, female) enjoyed reading aloud the most, comparing it to reading to children at daycare, and also enjoyed the robot's social commentaries, saying \textit{``it gives me a small little break from reading aloud and something to listen to besides me reading, for a minute.''} Over time, reading with the robot encouraged Alice to read more on her own, and helped her improve her pronunciations and confidence, \textit{``I think the robot makes me more conscious of myself reading.''} 
Furthermore, \textbf{Finn}'s (10yo, male) excitement and enjoyment towards the books extended to starting a shared experience with his family members, which was described by his parent as \textit{``we listened to one of those as an audio book when we were driving somewhere and our whole family listened to it and we really enjoyed it.''} Similarly, \textbf{Mona} (12yo, female) enjoyed the robot's companionship, saying \textit{``sometimes I tell my dad about the books that I'm reading, but it's nice to have someone else that's like saying comments about it, that actually knows the book.''} 
\textbf{Lynn} (10yo, female) enjoyed the experience of reading aloud to the robot, saying \textit{``the feedback that she gives about what’s going on makes me know that she’s listening, which makes me think that she probably likes reading with me.''} \textbf{Violet} (12yo, female) also enjoyed having \textit{``a partner who would make comments and listen''} and particularly liked the recall past interactions comments, \textit{``I really like to hear the ones where she refers to past books, and also like the ones where she just gets really excited about a subject or she'll just be enthusiastic about it.''} 

\section{Discussion and Conclusion}
In our four-week-long examination of children interacting with a learning companion robot, we see ample evidence that, in long-term real-world settings for human-robot interaction, it is challenging for children to follow a strict routine or prescribed schedule. In this study, we present findings on the key factors that emerged throughout the study that eventually dictated engagement with the robot and/or the activity. We found that there were \textit{domestic} factors including: differing levels of parental involvement, immediacy of the robot from the child's perspective, and disruptions to family routines that impacted engagement and interaction. We also found \textit{individual} factors about the child's perception of the robot and disposition toward reading and science, including their interest in the interaction routines with the robot, their perception of the social comments made by the robot, and their interest in the supplied reading materials. We believe that these factors can be translated into guidelines for the design of robots, particularly in anticipating factors that should be personalized to specific family and individual situations. We also believe that illustrating, through four cases, \textit{how} these factors influenced a wide variety of experiences and engagement with the robot builds our theoretical understanding of child-robot interaction.

\subsection{Key Factors for Practical Design in Long-Term Interaction}
In general, across these cases, it was rare for a child to find an ideal level of variety in the interaction or the reading materials, which suggests that single-activity oriented robot interactions may not effectively maintain engagement in the long-term. This finding is consistent with prior research that suggested that switching activities would maintain children's engagement with a social robot \cite{coninx2016towards}. Furthermore, while our interaction design followed guidelines presented by prior literature, \cite[e.g.,][]{gockley2005designing} regarding appearance, continuity and incremental behaviors, affective interactions and personality, and memory and adaptation, our findings demonstrate that for some children, i.e., who discontinued or adapted their interaction, there is still a gap for understanding the tailored design guidelines that are needed to sustain long-term interactions. In short, we have observed by our findings that a one-size-fits-all approach is not sustainable for long-term interactions which calls for a personalized and adaptive interaction design. Thus, we discuss how engagement in each case could have been improved by \textit{adaptive} strategies for responding to the key factors. Finally, our multi-modal approach to data collection reveals important nuances about interactions that cannot be captured by log data alone.

\paragraph{1. Multi-activity oriented long term robot interactions}
Our work involved a four-week deployment of a companion robot aimed to support children's reading. However, having one activity available for use was not ideal to keep children's engagement with the robot in the long term. We envision human-robot connections forming across a variety of activities, including but not limited to activities that support \textit{learning} (e.g., reading, assisting in math, or facilitating second language learning), \textit{entertainment and shared recreation} (e.g., playing board games, dancing, doing sports, playing music), or \textit{caretaking} of a robot (e.g., ``feeding'' the robot, getting the robot ready for bed, tucking the robot in at night). Activities that effectively achieve engagement can be identified through iterative development and testing, creating a modular design that supports the deployment of a set of activities based on the child's interests.

\paragraph{2. Adaptive strategies to address key factors that affect long-term robot interactions}

As in other long-term studies \cite[e.g.,][]{gockley2005designing}, several of our participants experienced periods of disruption preventing system use. For several users, individual interest in the reading materials and the robot system were sufficient motivators for continued use of the robot. However, other users lost interest due to a lack of interest in materials and a lack of novel robot behaviors---a factor from prior work that (negatively) affects long-term interactions \cite{leite2012modelling}. 
With careful design and testing, implementing the following adaptive strategies in response to short- and long-term interruptions can minimize the negative affects of loss in engagement: (1) connecting prior robot-interactions, (2) incorporating real life events into the interaction, and (3) connecting with the user about real life events. The robot can \textit{connect to prior interactions} by acknowledging when the last interaction occurred and expressing phrases aimed at repairing the connection with the user, such as \textit{``I've missed you,'' ``It's great to see you again,'' ``It's been a while since we last read,''} and checking in with the user by saying \textit{``How have you been,'' ``What are you up to these days,'' ``I hope everything is alright!''} Such expressions have the potential to enable the robot to reconnect with its user by showing that it cares and is sympathetic to the happenings in the user's external life. To \textit{incorporate real life events} into the interaction, after inquiring about the events in the user's personal life, the robot can engage in a brief conversation. For example, if the child expressed anticipation toward an event (e.g., a baseball game) and tells the robot about it, the robot can later ask how the event went. The robot can then \textit{connect with the user} by sharing its own past experiences about a similar event, for example, by saying \textit{``I've been to a baseball game before,''} and sharing memories and experiences with the child. We expect that building such connections early in the interaction will help form a stronger bond between the robot and the child and support sustained long-term interactions.

Children's varying level of individual interest in the interaction, the robot's comments, and provided books also limited engagement with the robot. Although we included 20 books over ten topics and implemented different variations in the robot's speech and commentary styles, our findings suggest that these variations were insufficient, possibly due to a lack of adaptive personalization--- a factor from prior work that positively affects long-term interactions \cite{leite2013social}. In long term deployments, robots should learn and adapt to children's personal preferences in interaction styles (e.g., task oriented, emotionally expressive, imaginative), adjust the types of comments it makes (e.g., as some children preferred informative comments over social comments), and adopt language used by the child through entertainment \cite{iio2015lexical}.

\paragraph{3. Multi-modal approach to data collection in long-term child-robot interactions}
Prior work in long-term interactions with social robots noted a need for more thorough robot logging in order to accurately determine user routines \cite{sung2009robots}. Our work accomplishes this goal by combining objective measures from the robot's interaction logs with subjective measures from interview data from the families. Without interviewing children about their experiences, we would not have identified contrasts between the interview and log data related to children's experiences. Our work also makes a methodological contribution to child-robot interactions, highlighting the importance of data collection beyond log and survey data. 

\subsection{Limitations and Future Work}

Our work was limited by several factors. First, due to COVID restrictions, the study was conducted fully-remotely as we provided a no-contact delivery of the resources to the participants and interacted with them over video calls. Our inability to setup the robot in-person resulted in minor protocol disruptions in the robot's first setup, which might have affected the first impressions some users had of the robot system and their subsequent usage. We also acknowledge that while our weekly interview protocol was necessary to maintain connections and build rapport with with families, it is likely that a number of participants' engagement with the robot was influenced by these weekly meetings as presented in our results. Furthermore, although the robot's interaction design included to have a level of adaptability and variety, our results show that children occasionally felt the robot was predictable and repetitive. This can be extended in future work by investigating the use of AI-based tools for narrative and comment generation to achieve such variability and integrate children's comments \textit{in situ}. Finally, our participant sampling was not balanced and limited due to the number of available robots, therefore we acknowledge that the findings presented through the data cannot be generalized.

In conclusion, we presented the interaction design of a learning companion robot aimed to support children's interest in reading by expressing social and informational comments in response to the book pages read by the children. Our four-week in-home deployment with fourteen families highlighted factors that impacted children's engagement with the robot, ranging from life events or ways of self-motivation to factors that stem from the robot's interaction. Findings from this in-home long-term child-robot interaction study contributes to exploring new approaches to improve future robot interaction designs for families.

\section{Selection and Participation of Children}
The study protocol was reviewed and approved by the University of Wisconsin-Madison Institutional Review Board (IRB). Children were recruited through their parents who were contacted through local community centers and university mailing lists. The inclusion criteria was families with at least one child aged 10--12 identified with low interest in science based on parent responses to a pre-screening survey. Non-eligible siblings were allowed to participate and interact with the robot if interested. For the consent process, researchers described the study to the family, obtained written consent from parents, and verbal assent from the minor(s). Children were encouraged to ask questions related to the study procedure and the study was initiated after the child clearly stated consent in participation. Parents received \$50 compensation on the last day of the study.

\begin{acks}
This work was supported by NSF DRL-1906854: ``Designing and Testing Companion Robots to Support Informal, In-home STEM Learning.''
\end{acks}

\bibliographystyle{ACM-Reference-Format}
\bibliography{sample-base}

 \end{document}

%% file: teaser.tex
\begin{figure*}[h!]
    \centering
  \includegraphics[height=1.8in]{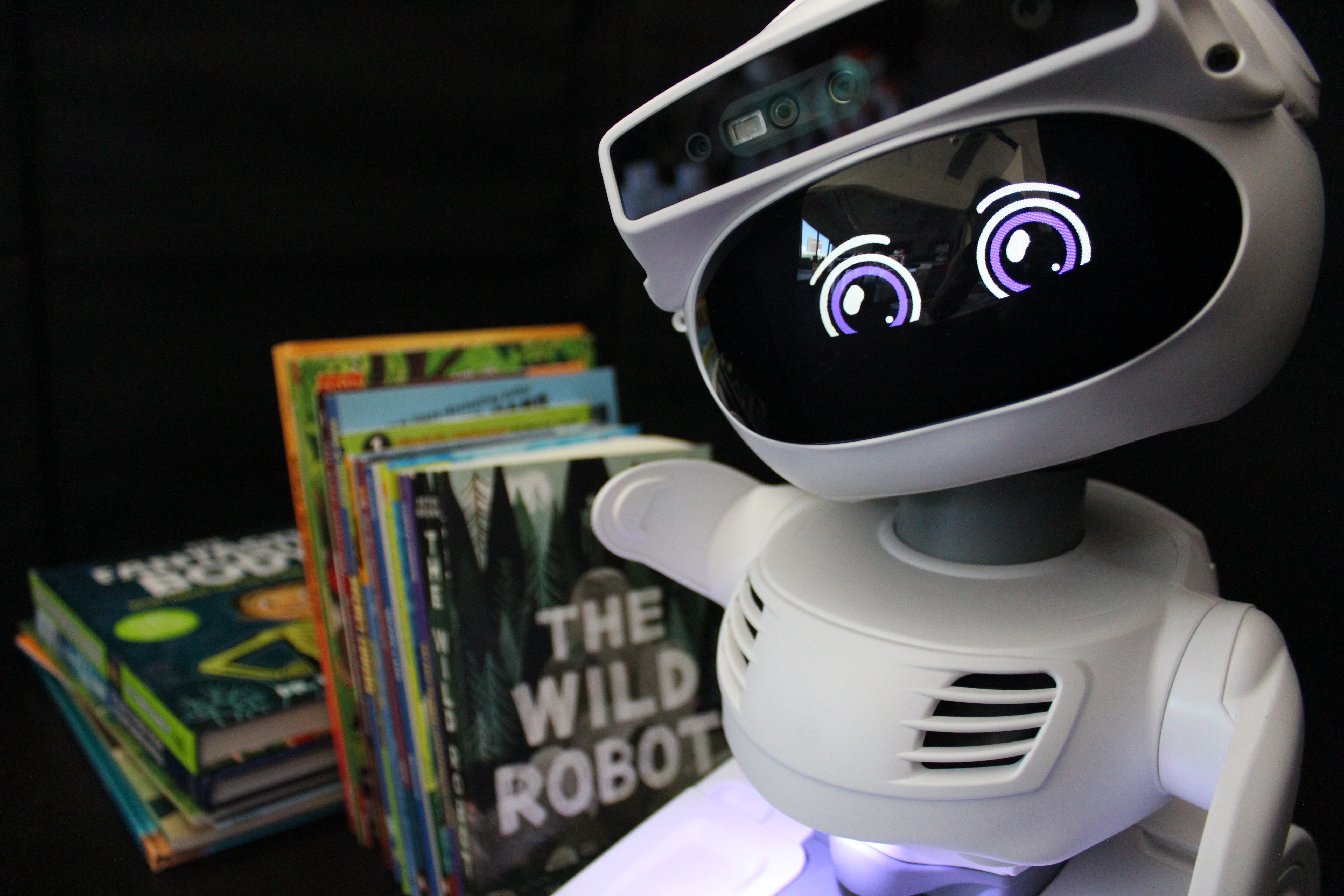}
    \includegraphics[height=1.8in]{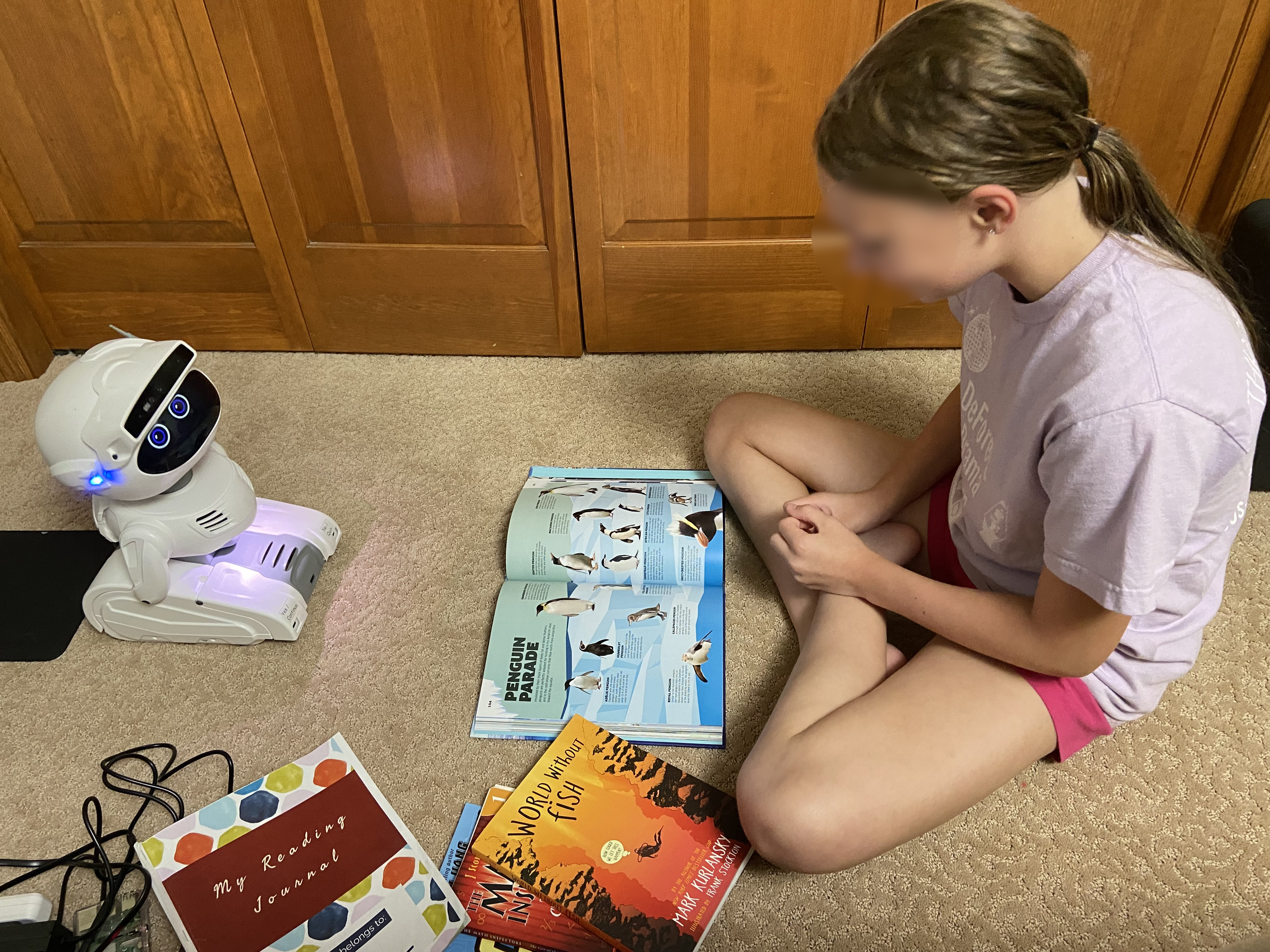}
  \caption{Our work explores the factors that shaped children's long-term engagement with a learning companion robot. Children read aloud informal science books to the robot for four weeks during the summer, in their own homes. The robot supported the reading experience by expressing knowledge, social, and interest comments as the child read to the robot. We report the factors that shaped children's reading habits and experiences with the robot and how it changed over time.}
  \label{fig:teaser}
  \vspace{-6pt}
\end{figure*}

%% file: fig_interactionflow.tex
\begin{figure*}[b!]
  \includegraphics[width=\linewidth]{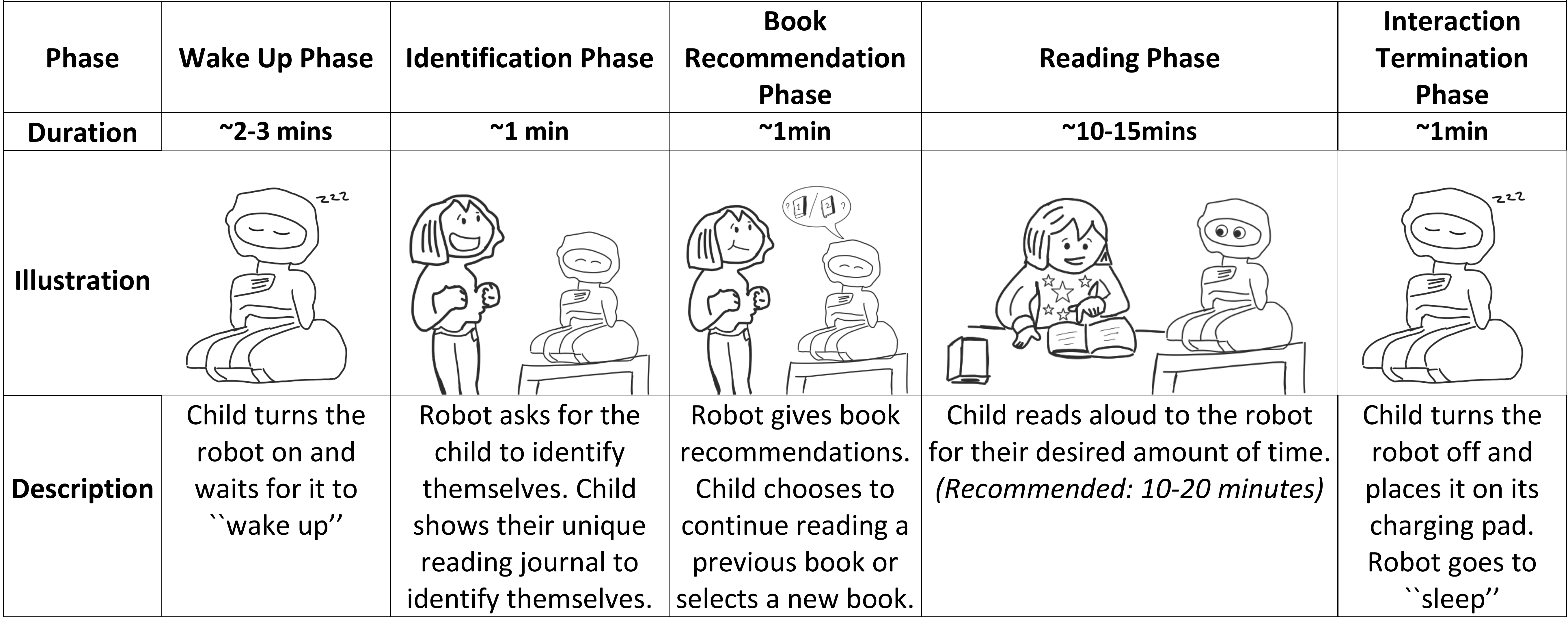}
  \caption{\textit{Descriptive illustration of the interaction flow with the robot:} The interaction flow consists of five phases, 1) wake up, 2) identification, 3) book recommendation, 4) reading, 5) interaction termination. A successful interaction with the robot is considered when a child goes through each of these phases, regardless of how long they read with the robot. 
  }
  \label{fig:flow}
\end{figure*}

%% file: fig_results.tex
\begin{figure*}[t!]
  \includegraphics[width=\linewidth]{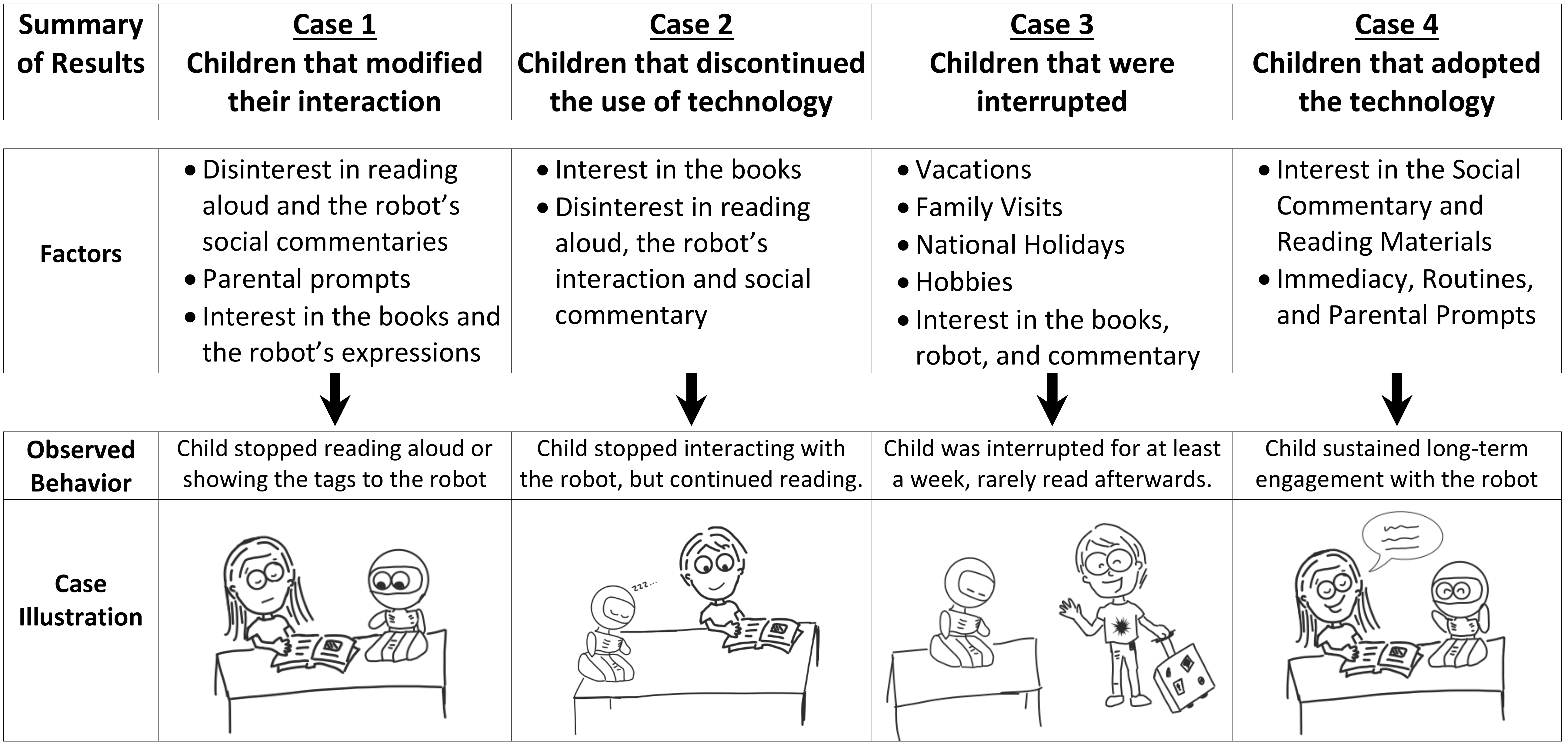}
  \caption{\textit{Summary of results:} We identified critical factors that impacted children's long term engagement with the robot. Based on children's observed behaviors derived from these critical factors, we categorized children in four cases: Children who 1) modified their interaction 2) discontinued the use of technology 3) were interrupted 4) adopted the technology. 
  }
  \label{fig:results}
\end{figure*}

%% file: fig_overallreading.tex
\begin{figure*}[t!]
  \includegraphics[width=2.8in]{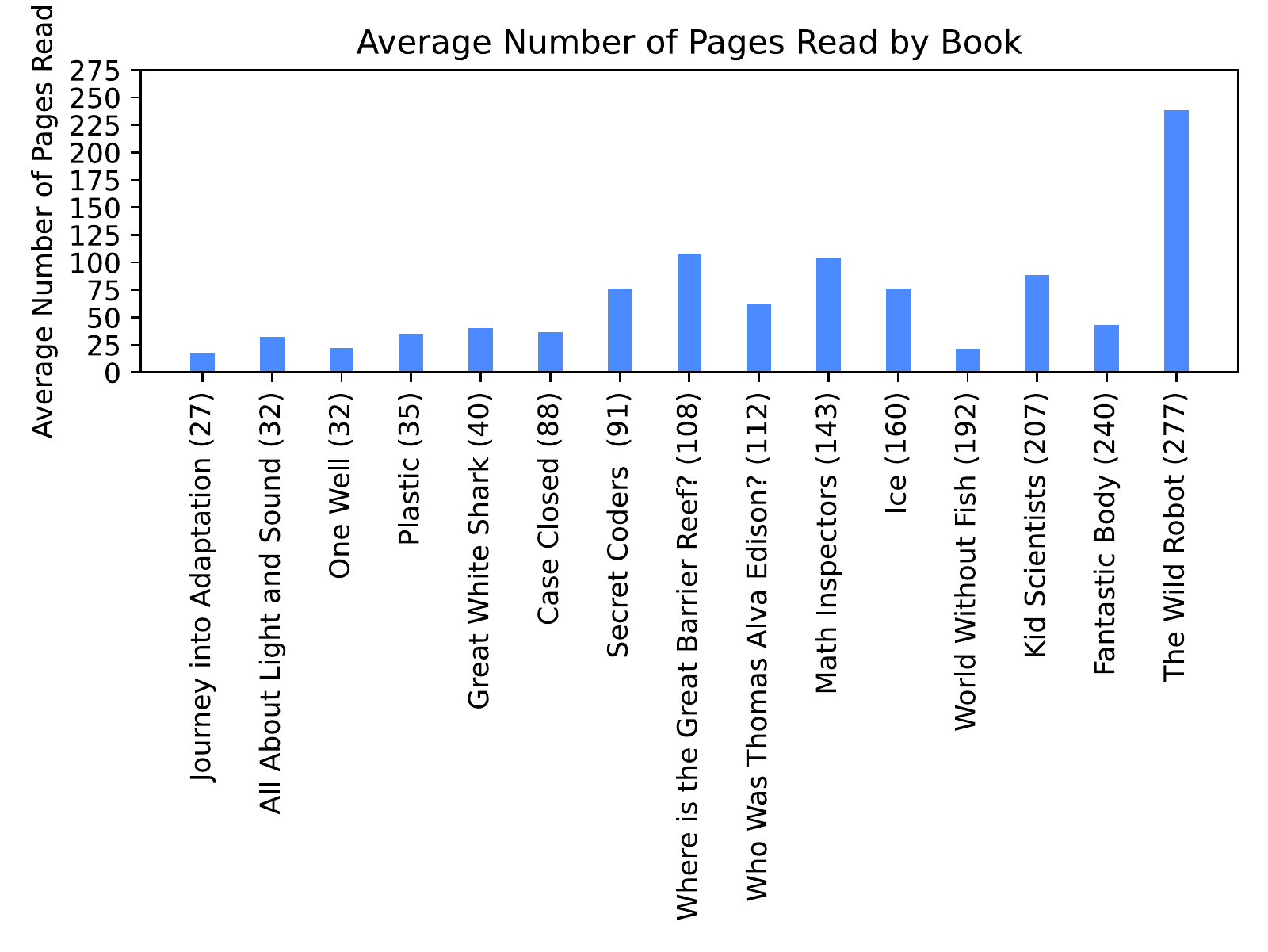}
    \includegraphics[width=2.8in]{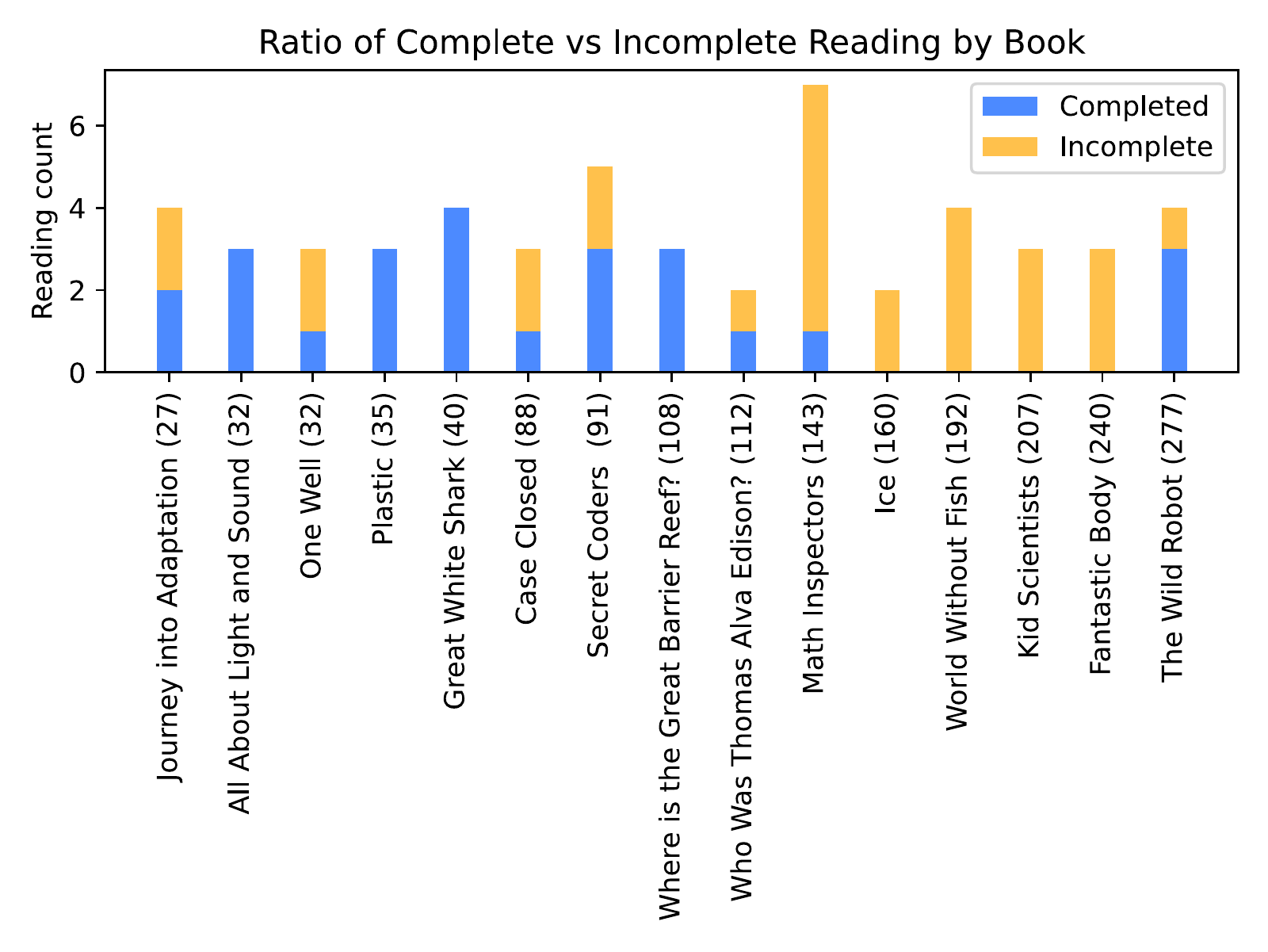}
  \caption{ Two graphs displaying children's \textit{overall} performance with the books throughout the study. The names of the books and their corresponding page numbers are presented in parenthesis. The graph on the (left) presents the percentage of pages read per book by all children. The graph on the (right) presents the number of times each book has been read by all children, and the ratio of the book's completion/incompletion rates.
  }
  \label{fig:booksread}
  \vspace{-6pt}
\end{figure*}